%% file: _main.tex
\input{_constants}
\arxiv

\pdfoutput=1
\documentclass[10pt,twocolumn,letterpaper]{article}
\input{cvpr_header}
\begin{document}
\title{Visual and Semantic Prompt Collaboration for Generalized Zero-Shot Learning}
\author{Huajie Jiang\textsuperscript{1}, Zhengxian Li\textsuperscript{1}, Xiaohan Yu\textsuperscript{2}, Yongli Hu\textsuperscript{1}\thanks{Corresponding author}, Baocai Yin\textsuperscript{1}, Jian Yang\textsuperscript{2}, Yuankai Qi\textsuperscript{2}\\ 
\textsuperscript{1}Beijing University of Technology  \ \ \textsuperscript{2}Macquarie University\\ 
{\tt\small \{jianghj,huyongli,ybc\}@bjut.edu.cn, lisin@emails.bjut.edu.cn,}\\ {\tt\small \{xiaohan.yu,jian.yang,yuankai.qi\}@mq.edu.au}
}
\maketitle

\input{00_abstract}
\input{01_intro}
\input{02_related}

\input{03_method}

\input{04_experiment}

\input{05_conclusion}
\input{Acknowledge}

{\small
\bibliographystyle{ieee_fullname}
\bibliography{11_references}
}

\ifarxiv \clearpage \input{12_appendix} \fi

\end{document}


\title{\paperTitle \\ Supplemental Material}
\author{\authorBlock}
\maketitle

\input{12_appendix}

{\small
\bibliographystyle{ieee_fullname}
\bibliography{11_references}
}

%% file: _constants.tex
\def\cvprPaperID{0000}
\def\confName{CVPR}
\def\confYear{2023}

\newif\ifreview 
\newif\ifarxiv \newcommand{\arxiv}{\arxivtrue}
\newif\ifcamera 
\newif\ifrebuttal 

%% file: cvpr_header.tex
\ifreview \usepackage[review]{cvpr} \fi
\ifarxiv \usepackage[pagenumbers]{cvpr} \fi
\ifrebuttal \usepackage[rebuttal]{cvpr} \fi
\ifcamera \usepackage{cvpr} \fi

\usepackage{graphicx}
\usepackage{amsmath}
\usepackage{amssymb}
\usepackage{booktabs}

\input{_macros}  

\usepackage{xr-hyper}

\makeatletter
\newcommand*{\addFileDependency}[1]{
  \typeout{(#1)}
  \@addtofilelist{#1}
  \IfFileExists{#1}{}{\typeout{No file #1.}}
}

\makeatother

\usepackage[pagebackref,breaklinks,colorlinks]{hyperref}
\usepackage[capitalize]{cleveref}
\crefname{section}{Sec.}{Secs.}
\crefname{table}{Table}{Tables}
\crefname{figure}{Fig.}{Figs.}

%% file: _macros.tex
\usepackage{times}
\usepackage{microtype}
\usepackage{epsfig}
\usepackage[table,xcdraw]{xcolor}
\usepackage{caption}
\usepackage{float}
\usepackage{placeins}
\usepackage{color, colortbl}
\usepackage{stfloats}
\usepackage{enumitem}
\usepackage{tabularx}
\usepackage{xstring}
\usepackage{multirow}
\usepackage{xspace}
\usepackage{url}
\usepackage{subcaption}
\usepackage{xcolor}
\usepackage[hang,flushmargin]{footmisc}

\ifcamera \usepackage[accsupp]{axessibility} \fi

% ...

\ifarxiv  \fi

\newcommand{\R}[1]{{%
    \textbf{%
        \ifstrequal{#1}{1}{\textcolor{red}{R#1}}{%
        \ifstrequal{#1}{2}{\textcolor{blue}{R#1}}{%
        \ifstrequal{#1}{3}{\textcolor{magenta}{R#1}}{%
        \ifstrequal{#1}{4}{\textcolor{teal}{R#1}}{%
                           \textcolor{cyan}{R#1}%
        }}}}%
    }%
}}

%% file: 00_abstract.tex
\begin{abstract}
Generalized zero-shot learning aims to recognize both seen and unseen classes with the help of semantic information that is shared among different classes.
It inevitably requires consistent visual-semantic alignment. 
Existing approaches fine-tune the visual backbone by seen-class data to obtain semantic-related visual features, which may cause overfitting on seen classes with a limited number of training images. 
This paper proposes a novel visual and semantic prompt collaboration framework, which utilizes prompt tuning techniques for efficient feature adaptation. 
Specifically, we design a visual prompt to integrate the visual information for discriminative feature learning and a semantic prompt to integrate the semantic formation for visual-semantic alignment. 
To achieve effective prompt information integration, we further design a weak prompt fusion mechanism for the shallow layers and a strong prompt fusion mechanism for the deep layers in the network. 
Through the collaboration of visual and semantic prompts, we can obtain discriminative semantic-related features for generalized zero-shot image recognition. 
Extensive experiments demonstrate that our framework consistently achieves favorable performance in both conventional zero-shot learning and generalized zero-shot learning benchmarks compared to other state-of-the-art methods.
\end{abstract}

%% file: 01_intro.tex
\section{Introduction}
\label{sec:intro}
Generalized Zero-Shot Learning (GZSL)~\cite{NIPS2009_1543843a} aims to recognize images from both seen and unseen categories with the help of class semantic information, where class attributes ~\cite{5206594,DBLP:conf/iccv/Jiang0SC19,10246362,xie2019attentive} and text descriptions ~\cite{Ba_2015_ICCV,reed2016learning,christensen2023image} are widely used to transfer knowledge from seen classes to novel classes. 
GZSL is intrinsically inspired by the human cognitive ability to understand unknown concepts \cite{5206594,larochelle2008zero,NIPS2009_1543843a}, gaining increasing attention recently.

\begin{figure}[t]
  \centering
   \includegraphics[width=1.0\linewidth]{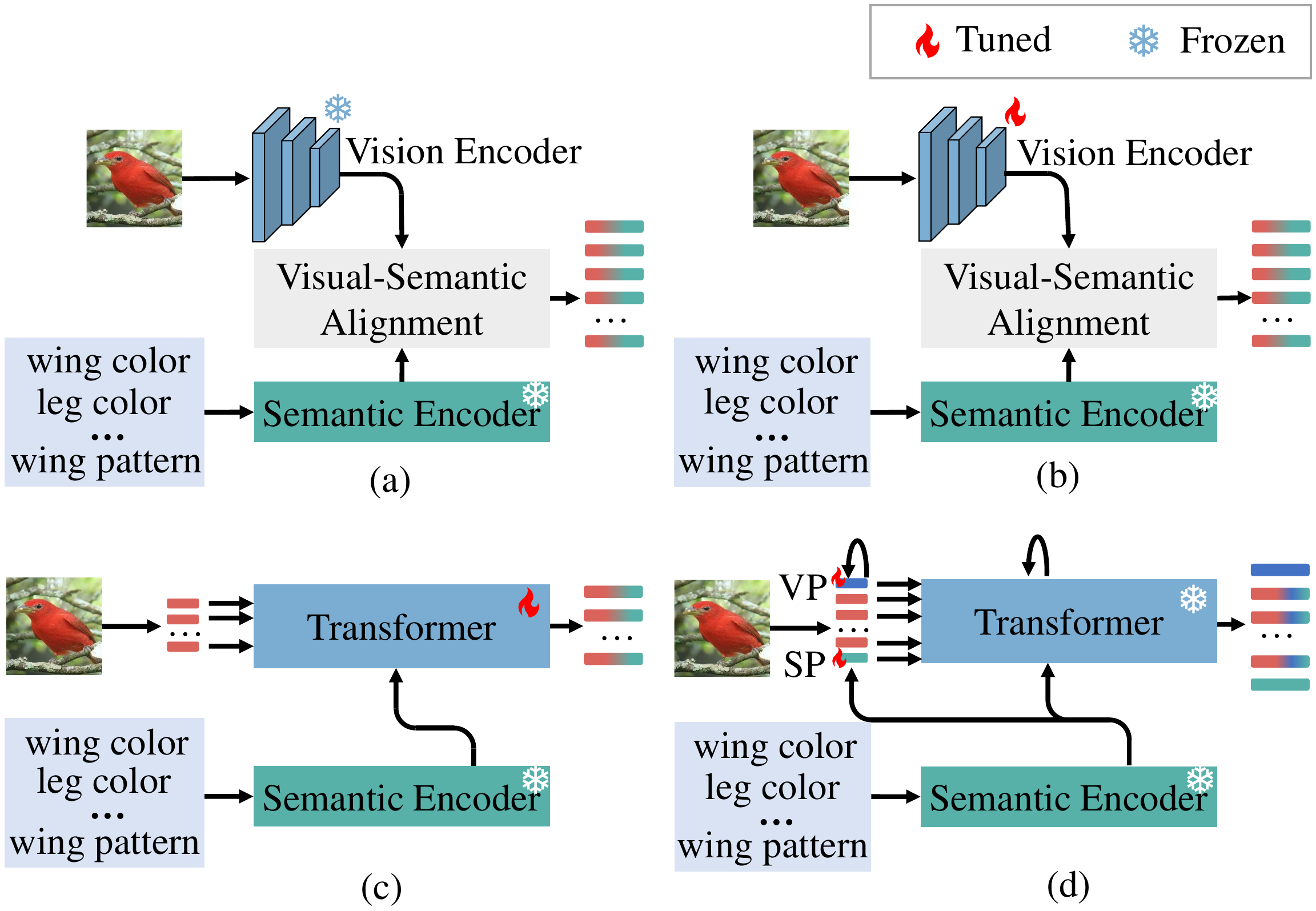}
   \caption{Different paradigms of GZSL. (a) Visual-semantic alignment with fixed backbone. (b) Fine-tuning visual features. (c) Fine-tuning with transformer backbone. (d) Our visual and semantic prompts collaboration network (VSPCN). The VSPCN integrates visual information and semantic information into the intermediate layers of the vision transformer network for semantic-related visual feature learning.}
   \label{fig:motivation}
   \vspace{-1.0em} 
\end{figure}

Traditional approaches \cite{DBLP:conf/cvpr/AkataPHS13,DBLP:conf/cvpr/MorgadoV17,DBLP:conf/cvpr/ZhangXG17,DBLP:conf/cvpr/0004YG18,DBLP:conf/cvpr/ReedALS16,DBLP:conf/eccv/JiangWSC18} extract the visual features and semantic features by pre-trained visual and language encoders independently and then align 
them in a common embedding space, 
as shown in Figure \ref{fig:motivation}(a). 
However, visual features obtained via pre-trained models may not be well-aligned with the semantic features since they are independently extracted.
To deal with this problem, some approaches \cite{xie2019attentive,zhu2019semantic,xu2020attribute,chen2023duet} propose to fine-tune the visual backbone to align with
class semantics, as shown in Figure \ref{fig:motivation}(b). 
Despite progress, the visual-semantic alignment is performed in the last layer of the network, which has less influence on the previous layers, resulting in less effective visual-semantic interaction. 
With the rapid development of the visual backbone network, transformer-based approaches have been proposed in zero-shot learning \cite{Zhao2023M3RMT, liu2023progressive, chen2024progressive}. They perform visual-semantic interaction for different layers in the model, enabling more semantic-related features, as is shown in Figure \ref{fig:motivation}(c).
However, these approaches tend to overfit seen classes in the fine-tuning process due to limited seen-class data.

To tackle this problem, we propose to utilize the prompt tuning technique to efficiently adapt the pre-trained model, and design a 
visual and semantic prompt collaboration network (VSPCN) for GZSL.
Different from traditional prompt learning approaches, which only learn visual prompts, our VSPCN learns both visual and semantic prompts for better adaptive visual feature learning, as shown in Figure \ref{fig:motivation}(d). 
The visual prompt aims to learn discriminative information by interacting with visual features, and the semantic prompt aims to learn class semantic information by interacting with the attributes. 
To further achieve effective interaction of prompts, visual features, and attributes, we design two types of information fusion mechanisms: weak prompt fusion and strong prompt fusion, which are incorporated into the shallow and deep layers of our model.
Through the collaboration of visual and semantic prompts, we can obtain more effective semantic-related visual features. 
Simple yet effective, our method achieves the best performance on three benchmark GZSL datasets.

To summarize, our main contributions are as follows:
\begin{itemize}
    \item We propose a visual and semantic prompt collaboration network that facilitates discriminative and semantic-relevant visual feature learning for generalized zero-shot recognition. 
    \item  We design weak and strong prompt fusion modules tailored to different model layers, achieving effective information fusion among prompts, visual features, and semantic features.
    \item Our method achieves the best performance on three GZSL benchmark datasets compared to other state-of-the-art approaches, demonstrating its effectiveness.
\end{itemize}

\begin{figure*}
  \centering
  \includegraphics[width=1.0\linewidth]{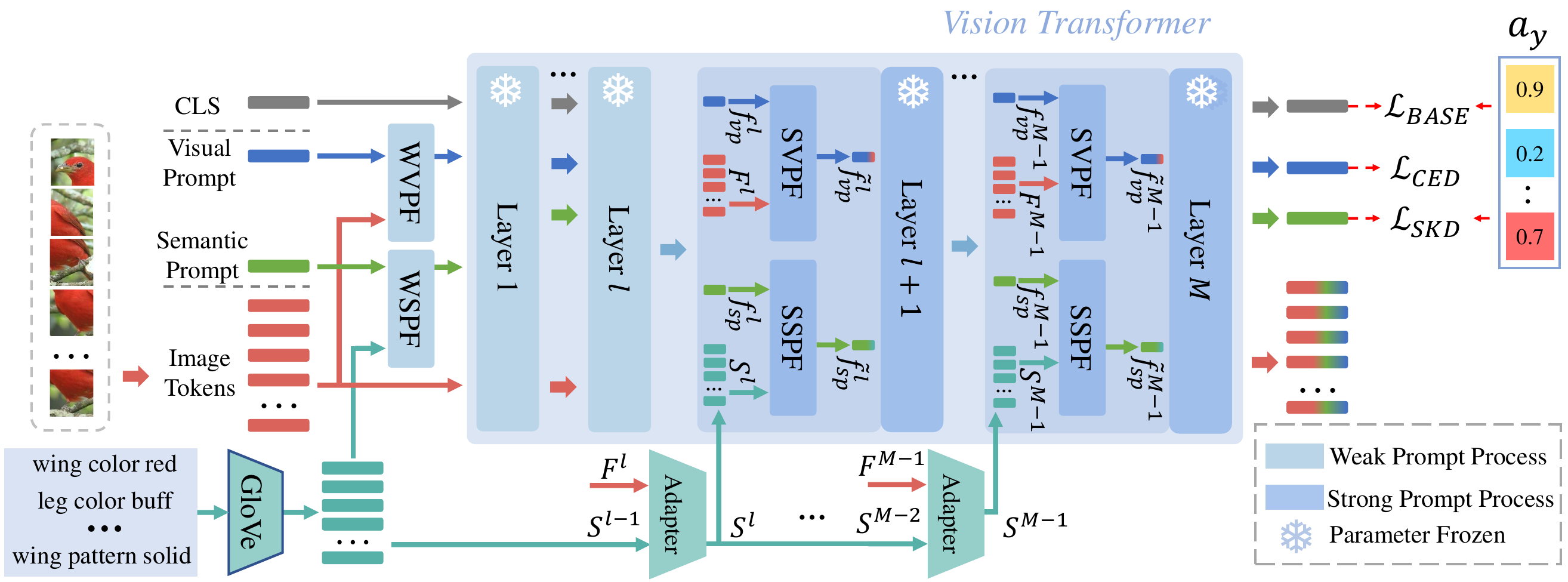}
  \caption{The framework of visual semantic prompts collaboration network (VSPCN). VSPCN utilizes the collaboration of visual and semantic prompts to adapt the pre-trained ViT model to the GZSL task. 
  The visual prompt learns to extract visual information from image tokens, and the semantic prompt incorporates semantic information from semantic attributes.
 Weak prompt fusion (including weak visual prompt fusion (WVPF) and weak semantic prompt fusion (WSPF)) (\ref{Weak Prompts Collaboration}) and strong prompt fusion (including strong visual prompt fusion (SVPF) and strong semantic prompt fusion (SSPF)) (\ref{Strong Prompts Collaboration}) mechanisms are designed to integrate visual and semantic information.
 Furthermore, the adapters are utilized to update the semantic features for instance-level adaptive semantic information extraction (\ref{Model Optimization and Inference}).}
  \label{fig:framwork}
\vspace{-1.0em} 
\end{figure*}

\begin{figure}[t]
  \centering
   \includegraphics[width=1.0\linewidth]{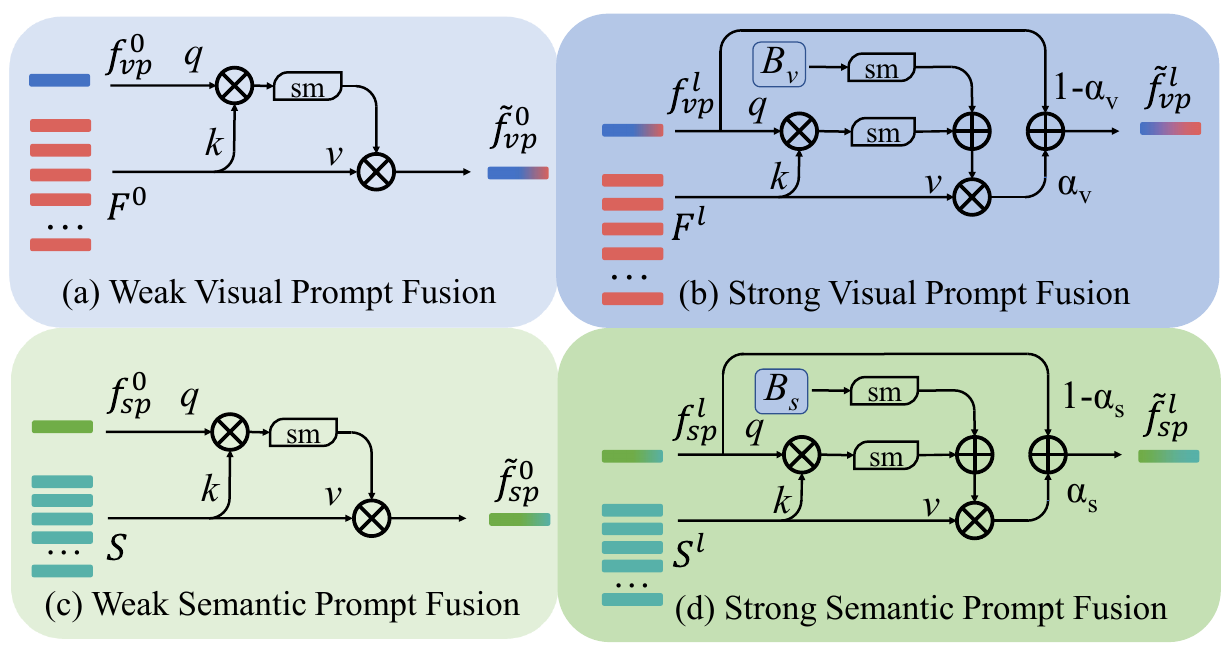}
   \caption{Illustration of our fusion modules. (a) weak visual prompt fusion, (b) strong visual prompt fusion, (c) weak semantic prompt fusion, and (d) strong semantic prompt fusion.}
   \label{fig:fusion}
\vspace{-1.0em} 
\end{figure}

%% file: 02_related.tex
\section{Related Work}
\label{sec:formatting}

\noindent\textbf{Generalized Zero-Shot Learning.}
Semantic information plays an important role in GZSL. Based on the approach of utilizing semantic information, current GZSL methods can be roughly divided into two categories: generative-based GZSL methods and embedding-based GZSL methods. 

Generative-based GZSL approaches aim to generate image features for the unseen classes using different generative models, such as generative adversarial nets (GANs) \cite{xian2018feature,8953480,9879288,chen2023deconstructed,gupta2023generative, 10400439,hou2024visual,10706204,DBLP:conf/eccv/JiangWSC18}, variational autoencoders (VAEs) \cite{narayan2020latent,verma2018generalized,9879149,10543099,zhang2024adaptive}, and denoising diffusion models \cite{10415463,nava2022meta,clark2024text}. Then they train GZSL classifiers using both the real seen-class visual features and the synthesized unseen-class visual features for recognition. Although these approaches have achieved great progress, the generative models may be difficult to train. Moreover, the feature-generation process is independent of the recognition process, which makes the generated features less effective for recognition.

The embedding-based GZSL methods aim to align visual space and semantic space through visual-semantic mapping and perform image recognition in a common semantic space. Earlier embedding-based works ~\cite{song2018transductive, xian2016latent, DBLP:conf/cvpr/ZhangXG17,8715419,8237715} directly project the global visual features into the sharing space for category predictions. However, the global visual features fail to capture local discriminative representations, leading to sub-optimal classification performance. Therefore, the attention-based approaches~\cite{xie2019attentive, xu2020attribute, chen2022transzero, chen2022msdn}, have been adopted to highlight discriminative visual features related to attributes. Recent works \cite{wang2021dual,liu2023progressive,chen2024progressive,zhou2024dynamic} deploy the semantic-visual attention module to progressively update semantic attributes and visual features together, which improves the consistency of visual features and semantic features. Though great progress has been made, these works fine-tune the backbone with limited numbers of seen-class data, improving the risk of overfitting the seen classes. 

Different from existing approaches, which fine-tune the backbone to obtain semantic-relevant visual features, we adopt an effective prompt learning mechanism to adapt the pre-trained model to the GZSL task.

\noindent\textbf{Prompt Learning for Vision Transformer.}
Prompt learning bridges the gap between pre-training models and downstream tasks, enabling effective adaptation to the downstream tasks. Vision transformers utilizing prompt engineering demonstrate remarkable performances in various visual tasks, such as compositional zero-shot learning \cite{nayak2022learning,lu2023decomposed,wang2023hierarchical}, few-shot learning \cite{chen2023semantic}, and continuous learning \cite{wang2022learning,wang2022dualprompt}. DFSP \cite{lu2023decomposed} and CSP \cite{nayak2022learning}
introduce soft prompt, a parameter-efficient learning technique that tunes a subset of tokens within the vocabulary to represent primitive
concepts. SP \cite{chen2023semantic} proposes a novel semantic prompt approach to leveraging textual information in class names for few-shot image recognition, which aims to adaptively adjust pre-trained visual features to class-specific features. L2P \cite{wang2022learning} and Dualprompt \cite{wang2022dualprompt} utilize prompt learning to learn new classes and simultaneously keep the knowledge of old classes. 

Different from the above prompt learning approaches, we propose a novel visual and semantic prompt collaboration framework, which simultaneously learns visual and semantic prompts for effective semantic-related visual feature learning. Moreover, we design weak and strong prompt fusion methods tailored to different ViT layers, achieving effective information fusion among prompts, visual features, and semantic features.

%% file: 03_method.tex
\section{Method}
\label{sec:method}

\noindent
\textbf{Problem Formalization.} Zero-Shot learning aims to transfer knowledge from seen class $\mathcal{D}^{s}$, to unseen classes $\mathcal{D}^{u}$ in the inference. $\mathcal{D}^{s}=\{(x^s,y^s,a_y^s)|x\in\mathcal{X}^s,y\in\mathcal{Y}^s,a_y^s\in\mathcal{A}^s\}$, where $x^s$ refers to the image in $\mathcal{X}^s$, $y^s$ refers to the corresponding label and $a_y^s$ refers to the corresponding category semantic attributes from seen class datasets. Similarly, $\mathcal{D}^{u}=\{(x^u,y^u,a_y^u)|x\in\mathcal{X}^u,y\in\mathcal{Y}^u,a_y^u\in\mathcal{A}^u\}$, where $x^u$ refers to an image in $\mathcal{X}^u$, $y^u$ refers to the corresponding label and $a_y^u$ refers to the corresponding category semantic attributes from unseen class datasets. Based on ZSL setting $\mathcal{Y}^s\cap\mathcal{Y}^u=\emptyset$, ZSL aims to learn a classifier for unseen classes $f_{zsl}:\mathcal{X}{\rightarrow \mathcal{Y}^u}$
predict the class labels $\mathcal{Y}$ from the unseen set $\mathcal{Y}^u$, while GZSL aims to predict the class labels $\mathcal{Y}$ = $\mathcal{Y}^u\cup\mathcal{Y}^u$ from both seen and unseen classes. In this paper, we use $S\in \mathbb{R}^{N_a\times D}$ to represent the sharing semantic attribute to describe the word vectors of attributes, which is abstracted by a language model \textbf{Glove} \cite{pennington2014glove}. Among them, $N_a$ and $D$ represent the number of attributes and $D$ represents the dimension of the Glove feature.

\noindent
\textbf{Overview.} In order to extract semantic-related features effectively, we propose a visual and semantic prompt collaboration Network (VSPCN), which utilizes the prompt learning mechanism to adapt the pre-trained model to specific GZSL tasks. Figure \ref{fig:framwork} provides an overview of the proposed method, where the pre-trained Vision Transformer (ViT) serves as the backbone network. VSPCN expects five inputs: CLS token, visual prompt, semantic prompt, image tokens, and shared semantic attributes. Among them, the CLS token, visual prompt, and semantic prompt are randomly initialized, image features are mapped from image patches, and shared semantic attributes are obtained from \textbf{Glove} \cite{pennington2014glove}. To ensure effective information fusion among prompts, image features, and attributes, we design weak prompt fusion in the shallow layer (Section \ref{Weak Prompts Collaboration}) and strong prompt fusion in the deeper layers (Section \ref{Strong Prompts Collaboration}).
Finally, we propose an entropy-based divergence loss for the visual prompt and a semantic knowledge distillation loss for the semantic prompt, ensuring discriminative and semantic feature learning (Section \ref{Model Optimization and Inference}).

\subsection{Weak Prompts Fusion}
\label{Weak Prompts Collaboration}
Weak prompt fusion aims to achieve information fusion in the early stage, which contains weak visual prompt fusion and weak semantic prompt fusion. As shown in Figure \ref{fig:fusion}(a), the weak visual prompt aims at fusing the local visual features $F^0=[f_1^0,f_2^0,...,f_{N_v}^0]\in{\mathbb{R}^{N_v\times D }}$ in each image token to extract visual prompt information, thereby enriching the randomly initialized visual prompt $f_{vp}^0$. Specifically, we utilize the attention mechanism to achieve information fusion, which can be formulated as:
\begin{equation}
    Q_v^0 = q(f_{vp}^0), 
    K_v^0 = k(F^0), 
    V_v^0 = v(F^0)
\end{equation}
\begin{equation}
    \tilde{f}_{vp}^0=\delta(\frac{Q_v^0{K_v^{0}}^T}{\sqrt{D}})V_v^0
\end{equation}
where $\tilde{f}_{vp}^0$ is a weak visual prompt updated by visual features, $T$ denotes the transpose function, $q(\cdot)$, $k(\cdot)$ and $v(\cdot)$ are the linear mapping functions for the query, the key and the value. The $\delta()$ is $softmax()$ operation. Similarly, in order to enhance the semantic prompts, we fuse semantic information from shared semantic attributes by semantic attention, as shown in Fig \ref{fig:fusion}(c):
\begin{equation}
    Q_s^0 = q(f_{sp}^0), 
    K_s^0 = k(S), 
    V_s^0 = v(S)
\end{equation}
\begin{equation}
    \tilde{f}_{sp}^0=\delta(\frac{Q_s^0{K_s^{0}}^T}{\sqrt{D}})V_s^0
\end{equation}
where $\tilde{f}_{sp}^0$ is a weak semantic prompt updated by fine-grained attribute features. Compared to the original semantic prompt $f_{sp}^0$, the weak semantic prompt $\tilde{f}_{sp}^0$ integrates semantic information and is closely linked to specific attribute features.

Weak prompt fusion is only conducted for the inputs. Then, we concatenate the CLS token, weak visual prompt, weak semantic prompt, and the image tokens to obtain $\tilde{F}^0=[f_{cls}^0,\tilde{f}_{vp}^0,\tilde{f}_{sp}^0,f_1^0,f_2^0,...,f_{N_v}^0]$, which is utilized to perform feature learning in subsequent transformer layer. Afterward, the visual prompt, the semantic prompt, and visual features will interact in the ViT backbone by multi-head attention, achieving adaptive information fusion.

\subsection{Strong Prompts Fusion}
\label{Strong Prompts Collaboration}
After calculating the first $l$ layers of ViT, we can obtain the patch $F^l=[f_{cls}^l,f_{vp}^l,f_{sp}^l,f_1^l,f_2^l,...,f_{N_v}^l]=[f_{cls}^l,f_{vp}^l,f_{sp}^l, F^l]$. Since semantic information is only utilized for the inputs to enhance the semantic prompt, the semantic influence will be weakened in deeper layers, so we need to enrich the visual and semantic prompt information for the deeper layers. For this purpose, we design a strong prompt fusion mechanism for visual prompts and semantic prompts. As shown in Figure \ref{fig:fusion}(b), we utilize an attention mechanism with bias estimation to learn the residual information to update the visual prompt $f_{vp}^l$. During the fusion process, the visual prompt is updated by image tokens and will not be affected by other tokens. When updating the visual prompt, the predicted attention bias $B_v^l \in \mathbb{R}^{N_v}$ are added to the attention matrix:
\begin{equation}
    Q_v^l = q(f_{vp}^l), 
    K_v^l = k(F^l), 
    V_v^l = v(F^l)
\end{equation}
\vspace{-8mm} 
\begin{align}
        \tilde{f}_{vp}^l= & [\alpha_v \delta(\frac{Q_v^l{K_v^l}^T}{\sqrt{D}})+(1-\alpha_v) \delta(B_v^l)]V_v^l  + f_{vp}^l
    \label{Eq:alpha_v}
\end{align}
where $l$ indicates layer number, $\alpha_v \in (0,1)$  represents the visual weight coefficient, $F^l$ is the visual tokens output of the previous layer $l$. To enrich the semantic information of the semantic prompts, we further introduce the semantic information from shared semantic attributes to update the semantic prompt helping ViT extract semantic-related visual features for zero-shot learning. As shown in Figure \ref{fig:fusion}(d), this fusion process of the semantic prompt is achieved by a transformer with attention bias $B_s^l$. The formulation of semantic prompt in strong fusion is as follows:
\begin{equation}
    Q_s^l=q(f_{sp}^l),
    K_s^l=k(S^l),
    V_s^l=v(S^l)
\end{equation}
\vspace{-5mm}
\begin{align}
    \tilde{f}_{sp}^l=&[\alpha_s \delta(\frac{Q_s^l{K_s^l}^T}{\sqrt{D}})+(1-\alpha_s)\delta(B_s^l)]V^l_s +f_{sp}^l
    \label{Eq:alpha_s}
\end{align}
where $S$ is sharing semantic attributes and is updated by the adapter under the influence of the image tokens, and $\alpha_s \in (0,1)$  represents the semantic weight coefficient. Therefore, the prior knowledge from pre-trained NLP models is better transferred to the ViT features through the semantic prompt.

After obtaining the enhanced visual prompt $\tilde{f}_{vp}^l$ and semantic prompt $\tilde{f}_{sp}^l$, we concatenate them with $[f_{cls}^l,f_1^l,f_2^l,...,f_{N_v}^l]$ to get $\tilde{F}^l = [f_{cls}^l,\tilde{f}_{vp}^l,\tilde{f}_{sp}^l,f_1^l,f_2^l,...,f_{N_v}^l]$, and then input $\tilde{F}^l$ to lay $l$+1 for fine-grained feature extraction. Similarly, the subsequent ViT layers undergo the same strong prompt fusion process of visual and semantic prompts.

To enhance the semantic information for consistent information fusion, we design an adapter to learn adaptive semantic information for effective strong prompt fusion. Specifically, we interact the semantic information with the image features to learn instance adaptive semantic features using an attention mechanism, which can be formulated as:

\begin{equation}
    Q_a^l = q(S^{l-1}),
    K_a^l = k(F^l), 
    V_a^l = v(F^l)
\end{equation}
\begin{equation}
    S^l=\alpha_a \delta(\frac{Q_a^l{K_a^l}^T}{\sqrt{D}})V_a^l+(1-\alpha_a) S^{l-1}
\end{equation}
where $\alpha_a \in (0,1)$  represents the hyperparmeter.
\subsection{Model Optimization and Inference}
\label{Model Optimization and Inference}
As shown in Fig.\ref{fig:framwork}, after completing the feature extraction of the ViT network, we obtain the final tokens $F^M = [f_{cls}^M,f_{vp}^M,f_{sp}^M,f_1^M,f_2^M,...,f_{N_v}^M]$. Then, we employ the base loss $\mathcal{L}_{BASE}$, the cross-entropy-based divergence loss $\mathcal{L}_{CED}$, and the semantic knowledge distillation loss $\mathcal{L}_{SKD}$ to optimize our VSPCN.

\noindent
\textbf{Base Loss.}
As a major component, the base loss $\mathcal{L}_{BASE}$ plays a role in image classification and semantic alignment and is composed of both the classification loss $\mathcal{L}_{CLS}$ and the semantic regression loss $\mathcal{L}_{AR}$. 
\begin{equation}
    \mathcal{L}_{BASE}=\mathcal{L}_{CLS}+\gamma\mathcal{L}_{AR}
\end{equation}
where $\gamma$ is the weight parameter. For the classification loss, we first calculate the similarity between CLS token $f_{cls}^M$ and the category semantic prototypes $\tilde{a}_y$, and then use the cross-entropy loss to calculate the classification loss:
\begin{equation}
    \begin{split}
        \tilde{a}_y=a_y\cdot W_d
    \end{split}
\end{equation}
\begin{equation}
    \begin{split}
        \mathcal{L}_{CLS}=-\log\frac{\exp(f_{cls}^M\cdot \tilde{a}_y^T)} {\sum\limits_{\hat{y}\in \mathcal{Y}^s}\exp(f_{cls}^M\cdot   \hat{a}_y^T)}
    \end{split}
\end{equation}
where $W_d$ denotes the embedding parameter with the size of $N_a\times D$. At the same time, to promote alignment between CLS tokens and their corresponding real semantic prototypes, we introduced the semantic regression loss to optimize VSPCN by minimizing the mean square error:
\begin{equation}
    \mathcal{L}_{AR}=||f_{cls}^M-\tilde{a}_{gt}||_2^2
\end{equation}

\noindent
\textbf{Cross-Entroy-Based Divergence Loss.}
In the process of ViT forward propagation, the visual prompt obtains prompt information from image tokens, so the visual prompt is similar to CLS tokens and can also be used for classification. However, to obtain divergence discriminative information different from CLS token, we propose a new regularization term, called the cross-entropy-based divergence loss $\mathcal{L}_{CED}$ consisting of the cross-entropy loss $\mathcal{L}_{CE}$ and the entropy-based divergence loss $\mathcal{L}_{ED}$.
\begin{equation}
    \mathcal{L}_{CED}=\eta_1\mathcal{L}_{CE}(\tilde{f}_{vp}^M\cdot W_c,y)+\eta_2\mathcal{L}_{ED}
\end{equation}
where $\eta_1$ and $\eta_2$ are the weights to control their corresponding loss terms, $y$ is the corresponding label and $W_c$ is the classifier with the size of $D\times N_s$ ($N_s$ denotes the number of seen classes). Meanwhile, to ensure the visual prompt $f_{vp}^M$ learns complementary information to the class token $f_{cls}^M$, we define the entropy-based divergence loss $\mathcal{L}_{ED}$:
\begin{equation}
    \begin{split}
        \mathcal{L}_{ED} = \log( \frac{\mathcal{L}_{CE}(f_{vp}^M\cdot W_c,y)+\mathcal{L}_{CE}(f_{cls}^M\cdot W_c,y)}{\mathcal{L}_{KL}(\delta(f_{vp}^M),\delta(f_{cls}^M))}+1)
    \end{split}
\end{equation}
where $\delta(\cdot)$ is a softmax function and $\mathcal{L}_{KL}$ is the Kullback-Leibler divergence loss \cite{1320776d-9e76-337e-a755-73010b6e4b64}, which guides the vision prompt to gain discriminative knowledge itself while complementary to the knowledge from the CLS token.

\noindent
\textbf{Semantic Knowledge Distillation Loss.}
To learn effective semantic prompts, we propose a new semantic knowledge distillation loss $\mathcal{L}_{SKD}$. It is composed of a Jensen-Shannon Divergence (JSD) and an Euclidean distance, which enables the semantic prompt to align with its corresponding class semantic prototype. Given the semantic prompt $f_{sp}^M$ and the semantic prototype $a_y$, semantic knowledge distillation loss $\mathcal{L}_{SKD}$ is defined as:
\begin{equation}
    \begin{split}
        \mathcal{L}_{SKD} = & \frac{1}{2}\mathcal{L}_{KL}(\delta(f_s^M),\delta(\tilde{a}_y))+\frac{1}{2}\mathcal{L}_{KL}(\delta(\tilde{a}_y),\delta(f_s^M)) \\
        & +||\tilde{a}_{gt}-f_s^M||_2^2
    \end{split}
\end{equation}

    To this end, we formulate the overall loss function of VSPCN as follows:
\begin{equation}                                        \mathcal{L}=\mathcal{L}_{BASE}+\lambda_{CED}\mathcal{L}_{CED}+\lambda_{SKD}\mathcal{L}_{SKD}
\label{Eq:loss}
\end{equation}
where $\lambda_{CED}$ and $\lambda_{SKD}$ are the hyper-parameters to control the importance of each loss. 

\noindent
\textbf{Inference}. After training VSPCN, we first obtain the CLS token of test instance $x$ through the ViT. Then, we apply an explicit calibration to adjust the predicted class probabilities and ensure balanced predictions across seen and unseen classes, which is formulated as follows:
\begin{equation}
 \tilde{y}=\arg\max\limits_{\hat{y}\in\mathcal{Y}^s\bigcup\mathcal{Y}^u}(f_{cls}^M\cdot a_{\hat{y}}^T+\tau\mathbb{I}_{\hat{y}\in \mathcal{Y}^u})
\end{equation}
where $\tau$ is a hyper-parameter to control the strength of the calibration applied to predictions for unseen classes, helping balance prediction biases between seen and unseen classes, and $\mathbb{I}_{\hat{y}\in \mathcal{Y}^u}$ is an indicator function which is zero when $\hat{y}\in\mathcal{Y}^s$.

%% file: 04_experiment.tex
\section{Experiment}

\begin{table*}
\begin{center}
\resizebox{\textwidth}{!}{
\begin{tabular}
{c|c|c|cccc|cccc|cccc}
\midrule
&\multirow{2}{*}{Methods} &\multirow{2}{*}{Backbone}&\multicolumn{4}{c|}{CUB} &\multicolumn{4}{c|}{SUN} &\multicolumn{4}{c}{AWA2}\cr
  & & & Acc & U & S & H & Acc & U & S & H & Acc & U & S & H\cr
\midrule
\midrule
\cellcolor{white}&LsrGAN(ECCV'20) \cite{vyas2020leveraging}  &ResNet-101 &60.3 &48.1 &59.1 &53.0    &62.5  &44.8 &37.7 &40.9  &66.4   &54.6  &74.6   &63.0  \\
\cellcolor{white}&CE-GZSL (CVPR'21) \cite{han2021contrastive} &ResNet-101  
&77.5 &63.9 &66.8 &65.3   &63.3  &48.8  &38.6 &43.1   &70.4   &63.1 &78.6 &70.0    \\
\cellcolor{white}&FREE (ICCV'21) \cite{chen2021free} &ResNet-101  
&- &55.7 &59.9 &57.7   &-  &47.4 &37.2 &41.7    &-  &60.4 &75.4 &67.1    \\
\cellcolor{white}&HSVA (NeurIPS'21) \cite{chen2021hsva}   &ResNet-101 &62.8 &52.7 &58.3 &55.3  &63.8  &48.6 &39.0 &43.3  &-  &56.7 &79.8 &66.3 \\
\multirow{-4}{*}{\cellcolor{white}\rotatebox[origin=c]{-90}{Generative}} &ICCE (CVPR'22) \cite{9879288}  &ResNet-101  &78.4  &67.3 &65.5 &66.4 &-  &- &- &- &72.7 &65.3 &82.3 &72.8    \\
\cellcolor{white}&VS-Boost (IJCAI'23) \cite{ijcai2023p123}   &ResNet-101 &- &68.0 &68.7 &68.4  &-  &49.2 &37.4 &42.5  &-  &- &- &- \\
\midrule
\cellcolor{white} &DAZLE (CVPR'20) \cite{9156868}&ResNet-101 &66.0 &56.7 &59.6 &58.1   &59.4  &52.3 &24.3 &33.2 &67.9  &60.3 &75.7 &67.1    \\
\cellcolor{white} &APN (NeurIPS'20) \cite{xu2020attribute}              &ResNet-101 &72.0 &65.3 &69.3 &67.2    &61.6 &41.9 &34.0 &37.6   &68.4 &56.5 &78.0 &65.5\\
\cellcolor{white} &GEM-ZSL (CVPR'21) \cite{liu2021goal}     &ResNet-101 &77.8 &64.8 &77.1 &70.4    &62.8 &38.1 &35.7 &36.9  &67.2 &64.8 &77.5 &70.6\\
\cellcolor{white} &DPPN (NeurIPS'21) \cite{wang2021dual} &ResNet-101  &- &70.2 &77.1 &73.5    &- &47.9 &35.8 &41.0     &- &63.1 &86.8 &73.1 \\
\cellcolor{white} &TransZero (AAAI'22) \cite{chen2022transzero}    
&ResNet-101 &76.8 &69.3 &68.3 &68.8    &65.6 &52.6 &33.4 &40.8  &70.1 &61.3 &82.3 &70.2\\
\cellcolor{white} &MSDN (CVPR'22) \cite{chen2022msdn}   &ResNet-101   &76.1 &68.7 &67.5 &68.1    &65.8 &52.2 &34.2 &41.3     &70.1 &62.0 &74.5 &67.7    \\
\cellcolor{white} &ICIS (ICCV'23) \cite{christensen2023image} &ResNet-101   &60.6 &45.8 &73.7 &56.5    &51.8 &45.2 &25.6 &32.7  &64.6 &35.6 &93.3 &51.6    \\
\cellcolor{white} &ZSCLR (ACM'24) \cite{yu2024zero} &ResNet-101   &77.8 &70.5 &74.3 &72.4    &66.3 &52.8 &45.3 &48.7  &\underline{76.0} &70.4 &76.7 &73.4    \\
\cellcolor{white} &ViT-ZSL (IMVIP'21) \cite{alamri2021multi} &ViT-Large  &- &67.3 &75.2 &71.0    &- &44.5 &55.3 &49.3     &- &51.9 &90.0 &68.5    \\
\cellcolor{white} &IEAM-ZSL (DGAM'21) \cite{alamri2021implicit} &ViT-Large &- &68.6 &73.8 &71.1    &- &48.2 &54.7 &51.3     &- &53.7 &89.9 &67.2    \\
\cellcolor{white} &DUET (AAAI'23) \cite{chen2023duet} &ViT-Base    &72.3 &62.9 &72.8 &67.5    &64.4 &45.7 &45.8 &45.8  &69.9 &63.7 &84.7 &72.7    \\
\cellcolor{white} &PSVMA (CVPR'23) \cite{liu2023progressive}&ViT-Base &- &70.1 &77.8  &\underline{73.8}   &-&61.7 &45.3  &\underline{52.3}  &-&73.6 &77.3 &\underline{75.4}\\
\cellcolor{white} &ZSLViT (CVPR'24) \cite{chen2024progressive} &ViT-Base      &\underline{78.9} &69.4 &78.2 &73.6    &\underline{68.3} &45.9 &48.4 &47.3    &70.7 &66.1 &84.6 &74.2\\
\multirow{-15}{*}{\cellcolor{white}\rotatebox[origin=c]{-90}{Embedding}} &VSPCN (Ours)  &ViT-Base      &\textbf{80.6} &72.8 &78.9 &\textbf{75.7}  &\textbf{75.3}  &59.4 &49.1 &\textbf{53.8}     &\textbf{76.6} &71.8 &84.3 &\textbf{77.6}    \\
\midrule
\end{tabular}
}
\end{center}
\vspace{-2.0em} 
\caption{Results on CUB, SUN, and AWA2, including generative-based methods and embedding-based methods. 'Acc' denotes the accuracy of unseen classes in the setting of CZSL, `U' denotes the accuracy of the unseen classes, `S' denotes the accuracy of the seen classes, and `H' is the harmonic mean in the setting of GZSL,  which is the main measurement for GZSL. `H' is the main metric for GZSL performance. `-' denotes that the result is not reported. The best and second-best results are marked in \textbf{}{blod} and \underline{underline}, respectively.}
\label{table:results comparison}
\vspace{-1.0em} 
\end{table*}

\subsection{Experimental Settings}
\noindent
\textbf{Benchmark Dataset.} In order to demonstrate the effectiveness of the proposed visual and semantic collaboration framework, we follow ZSLViT to perform evaluation on three datasets
Caltech-USCD Birds-200-2011 (CUB) \cite{Wah2011TheCB}, SUN Attribute (SUN) \cite{6247998} and Animals with Attributes2 (AWA2) \cite{xian2018zero}. We follow the split setting according to Proposed Split (PS) \cite{xian2018zero} to split each dataset into seen and unseen classes.

\noindent
\textbf{Evaluation Protocols.} During inference, we follow the evaluation protocol proposed by \cite{xian2018zero}. In the CZSL setting, we measure the accuracy of the test samples from the unseen classes (denoted  $Acc$). In the GZSL setting, we compute the accuracy of the test samples from both seen classes (denoted $S$) and unseen classes (denoted as $U$). To generally evaluate the performance of VSPC in the GZSL setting, we use the harmonic mean ($H$), which balances the seen class accuracy $S$ and the unseen class accuracy $U$. The harmonic mean is defined as $H=2\times S\times U/(S+U)$.

\noindent
\textbf{Implementation Details.} Following the PSVMA \cite{liu2023progressive}, we adopt the ViT-Base \cite{dosovitskiy2020image} as the feature extractor, which is pre-trained on ImageNet-1k. We set the parameter $l$=6 and perform weak prompt fusion in the first 6 layers of vit, followed by strong prompt fusion in the following layers. We use the Adam optimizer with hyper-parameters (learning rate=0.001, weight decay=0.0001) to optimize our model. We implement our approach with PyTorch on NVIDIA RTX A4000.

\subsection{Comparsion with State-of-the-Arts}
\noindent
\textbf{Conventional Zero-Shot Learning.}
Here, we first compare our VSPCN with the state-of-the-art methods in CZSL setting, where the performance is evaluated by the accuracy of the unseen classes in CZSL settings ($Acc$). As shown in Table \ref{table:results comparison}, our VSPCN achieves the best accuracies of 80.6\%, 75.3\%, and 76.6\% on three datasets compared with previous methods, including both embedding-based methods and generative ZSL methods. These results demonstrate the effectiveness of VSPCN for CZSL. In the CZSL scenario, compared to the second best methods (e.g., ZSLViT \cite{chen2024progressive} and ZSCLR \cite{yu2024zero}), our proposed method obtains gains over 1.7\%, 7.0\% and 0.6\% on CUB, SUN, and Awa2, respectively. This indicates that VSPCN can effectively learn semantic-related visual features adapted by dual prompts collaboration, which enhances the transferability from seen classes to unseen classes.

\noindent
\textbf{Generalized Zero-Shot Learning.}
Table \ref{table:results comparison} also presents the GZSL results of different methods, including CNN features-based and ViT features-based methods. In the GZSL setting, our VSPCN achieves the best harmonic mean $H$ of 75.7\%, 53.8\%, and 77.6\% on CUB, SUN, and Awa2, respectively.  Meanwhile, our VSPCN also outperforms the PSVMA \cite{liu2023progressive} by 2.9\%, 1.5\%, and 2.2\% on three datasets. 
In addition to harmonic mean $H$, we also perform well in the generalization of unseen classes, as can be seen by the $U$ measurement. These benefits are achieved by the collaboration of visual and semantic prompts in VSPCN, which effectively adapt the pre-trained ViT model to the GZSL task, enabling semantic-related features that effectively generalize for knowledge transfer from seen to unseen classes. Noted that even compared with  \cite{alamri2021multi,alamri2021implicit} whose backbone is trained on ImageNet-21k, our model pre-trained on ImageNet-1k still achieves better performance.

\subsection{Abalation Study}

\begin{table*}[!t]
\begin{center}
\resizebox{0.9\textwidth}{!}{
\begin{tabular}
{c|cc|cccc|ccc|ccc|ccc}
\midrule
\multirow{2}{*}{baseline}&\multirow{2}{*}{Pv} &\multirow{2}{*}{Ps}&\multirow{2}{*}{WVPF}&\multirow{2}{*}{WSPF}&\multirow{2}{*}{SVPF}&\multirow{2}{*}{SSPF}&\multicolumn{3}{c|}{CUB} &\multicolumn{3}{c|}{SUN} &\multicolumn{3}{c}{AWA2}\cr
  & & &&&& & U & S & H  & U & S & H  & U & S & H\cr
\midrule
\midrule
\checkmark\cellcolor{white} &&  & &&& &63.3 &55.9 &59.3      &67.6 &34.0 &45.2   &54.9 &79.9 &65.0\\
\checkmark\cellcolor{white}&\checkmark &  &\checkmark&& \checkmark 
& &68.7 &77.2 &72.7    &60.3  &44.8 &51.4       &60.5 &79.2 &68.6\\
\checkmark\cellcolor{white} &&\checkmark  & &\checkmark&&\color{blue}\checkmark &70.1 &78.0 &73.9      &59.2 &46.7 &52.2   &71.4 &81.7 &76.2\\
\midrule
\checkmark\cellcolor{white}   
&\checkmark &\checkmark&&&&&61.8 &70.0 &65.6    &56.6 &43.0 &48.9 &59.3 &77.4 &67.2\\
\checkmark &\checkmark  &\checkmark  &\checkmark&\checkmark&&  &73.5 &70.0 &71.7  &61.9 &45.6 &52.5    &70.1 &75.1 &72.5\\
\checkmark\cellcolor{white}&\checkmark  &\checkmark &&&\checkmark&\color{blue}\checkmark&73.6 &74.8 &74.2 &59.9  &45.6   &51.8  &68.1 &82.1 &74.4\\
\midrule
\checkmark\cellcolor{white} &\checkmark &\checkmark &\checkmark&\checkmark&\checkmark&\color{green}\checkmark  &72.5 &77.4 &74.9     &60.8 &46.9 &52.9       &65.7 &80.1 &72.2  \\
\checkmark\cellcolor{white} &\checkmark &\checkmark &\checkmark&\checkmark&\checkmark&\color{blue}\checkmark &72.8 &78.9 &\textbf{75.7}     &59.4 &49.1 &\textbf{53.8}     &71.8 &84.3 &\textbf{77.6}  \\

\midrule
\end{tabular}
}
\end{center}
\vspace{-1.5em} 
\caption{Ablation studies of different components in VSPCN. 'Pv' is the visual prompt, and 'Ps' means the semantic prompt. 'WVPF', ' WSPF', 'SVPF', and 'SSPF' denote the weak visual prompt fusion, weak semantic prompt fusion, strong visual prompt fusion, and strong semantic prompt fusion, respectively. The $\color{blue}\checkmark$ indicates that the semantic attributes have been updated by adapters, while $\color{green}\checkmark$ indicates no updation is conducted.}
\label{table:results ablation}
\vspace{-0.5em} 
\end{table*}

\begin{figure*}[!t]
    \centering
    \begin{minipage}{0.24\linewidth}
        \centering
        \includegraphics[width=\linewidth]{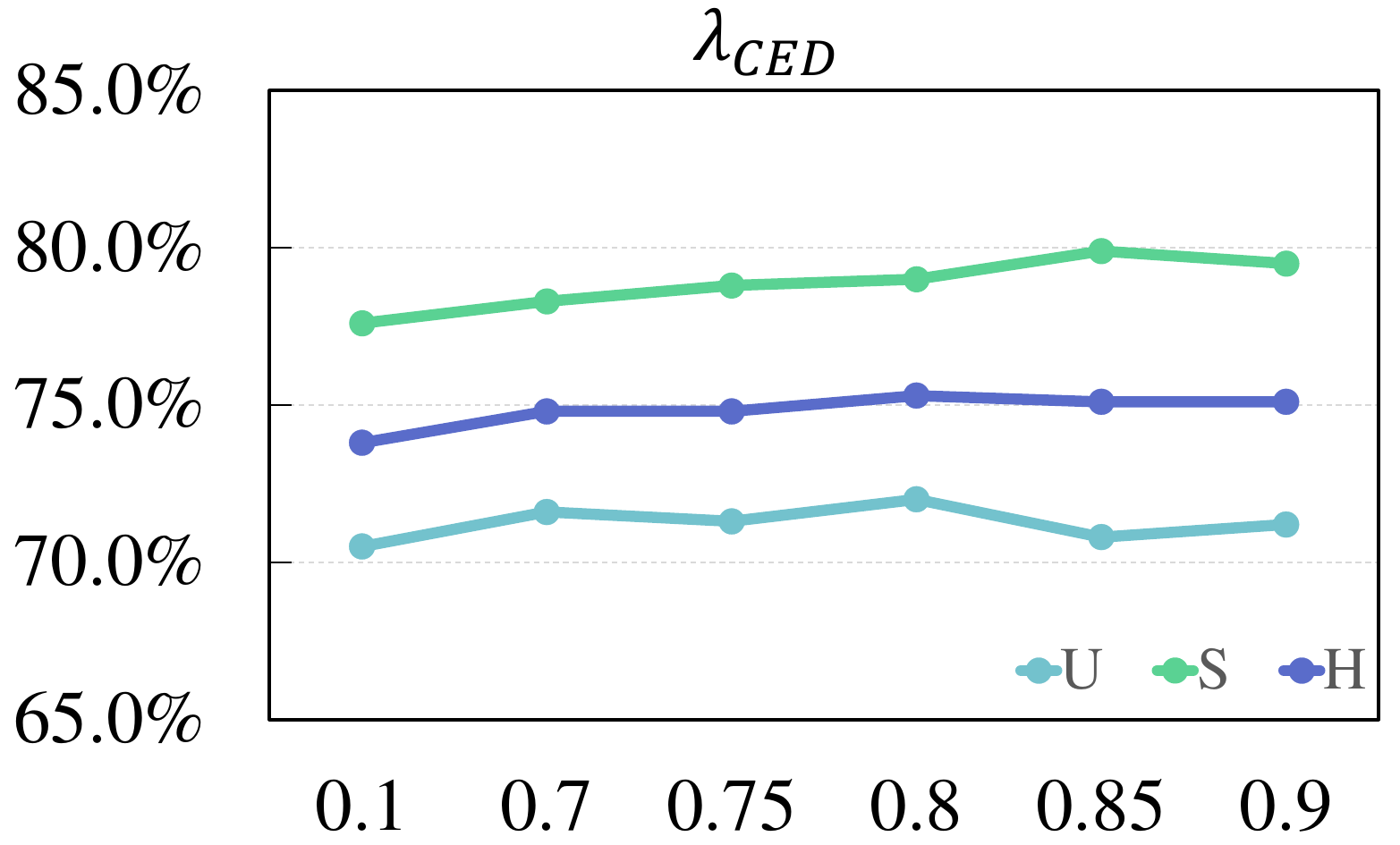}
        \captionsetup{labelformat=empty, skip=4pt}  
        \caption*{(a) CUB}
        \vspace{-1.0em} 
    \end{minipage}
    \hfill
    \begin{minipage}{0.24\linewidth}
        \centering
        \includegraphics[width=\linewidth]{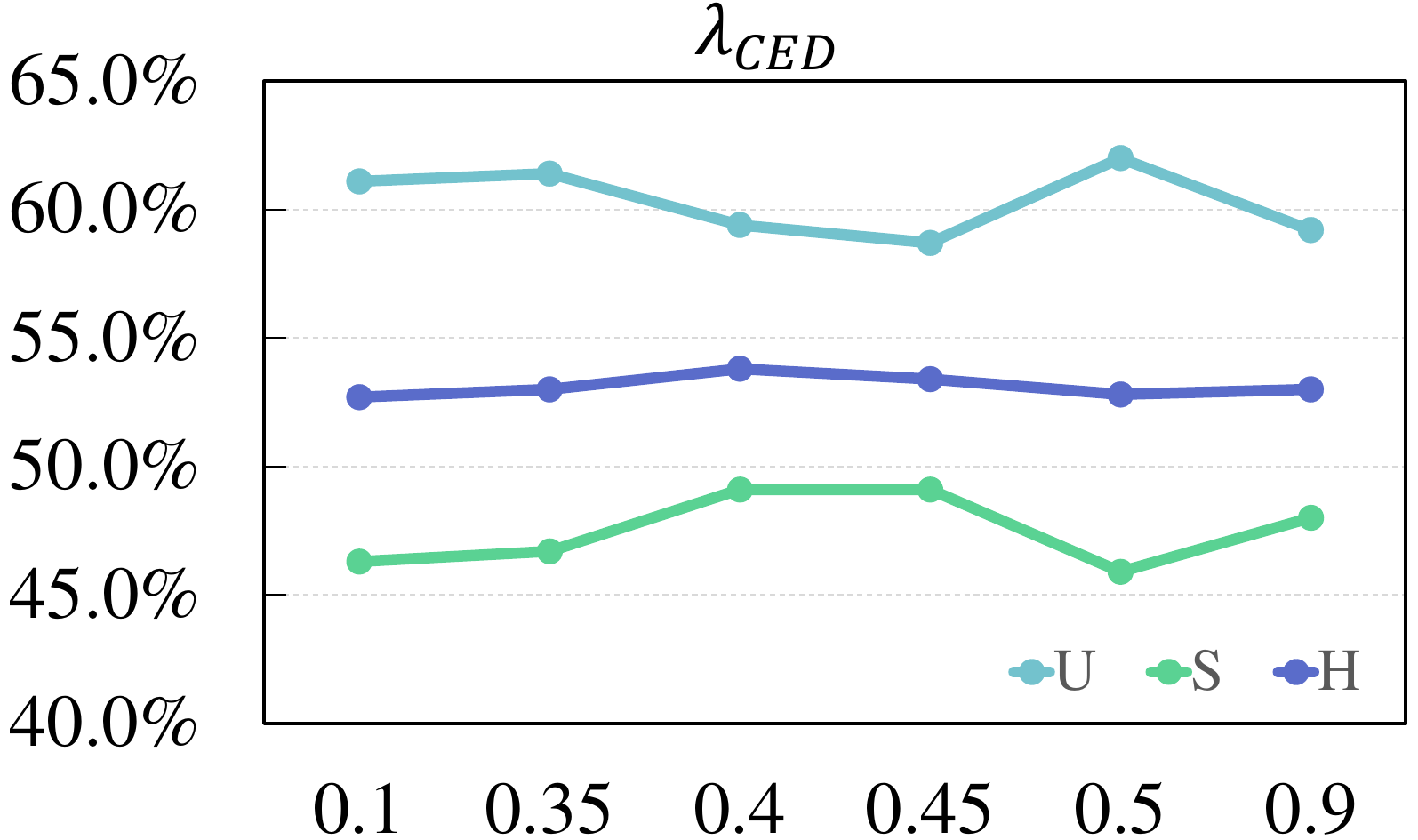}
        \captionsetup{labelformat=empty, skip=4pt}  
        \caption*{(b) SUN}
        \vspace{-1.0em} 
    \end{minipage}
    \hfill
    \begin{minipage}{0.24\linewidth}
        \centering
        \includegraphics[width=\linewidth]{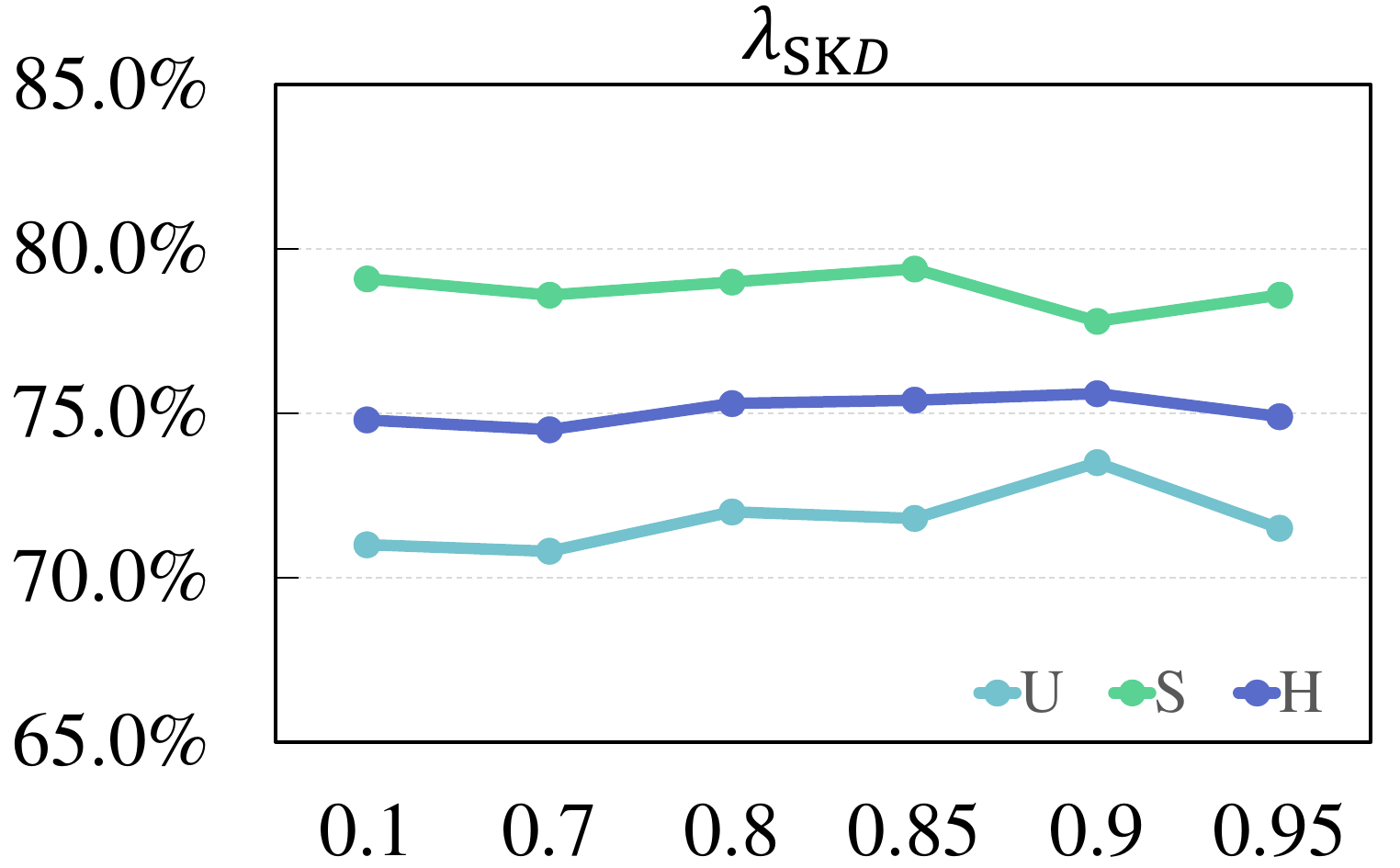}
        \captionsetup{labelformat=empty, skip=4pt}  
        \caption*{(c) CUB}
        \vspace{-1.0em} 
    \end{minipage}
    \hfill
    \begin{minipage}{0.24\linewidth}
        \centering
        \includegraphics[width=\linewidth]{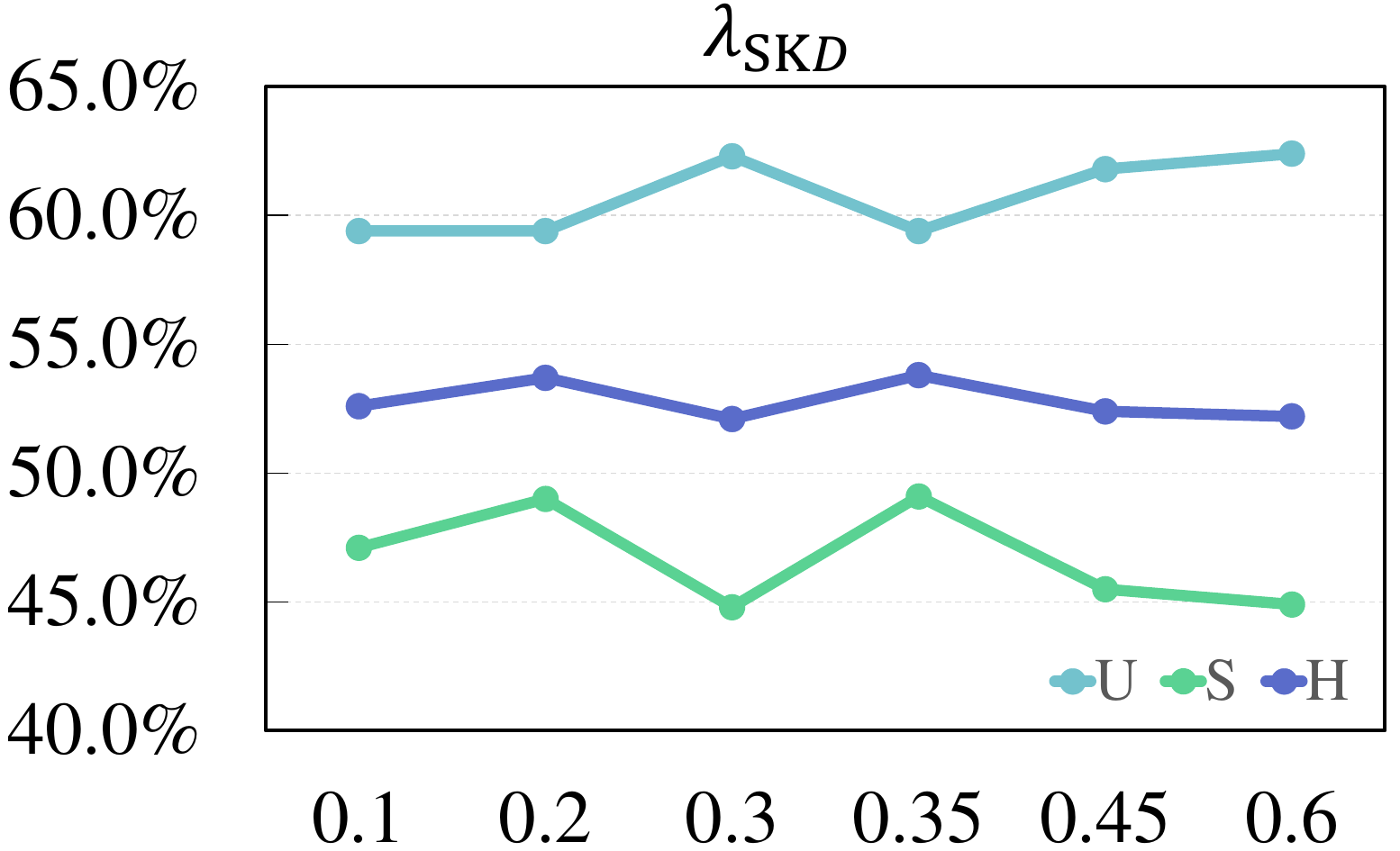}
        \captionsetup{labelformat=empty, skip=4pt}  
        \caption*{(d) SUN}
        \vspace{-1.0em} 
    \end{minipage}
\caption{The effects of hyperparameters $\lambda_{CED}$ and $\lambda_{SKD}$, which control the importance of visual and semantic prompts respectively.}
\vspace{-0.5em}
\label{fig:CED_SKD}
\end{figure*}

\begin{figure}[!t]
    \centering
    \begin{minipage}{0.48\linewidth}
        \centering
        \includegraphics[width=\linewidth]{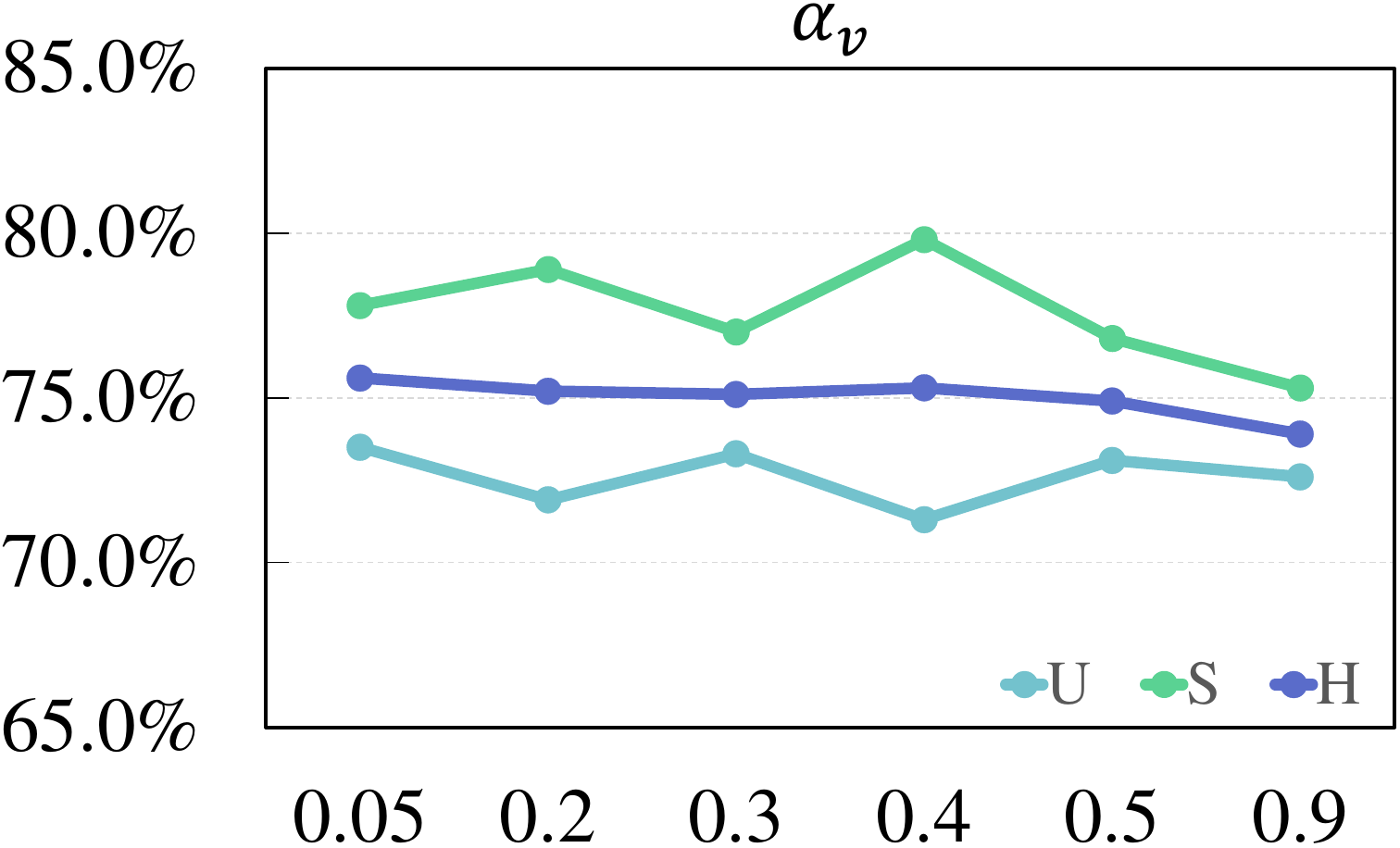}
        \captionsetup{labelformat=empty, skip=4pt}  
        \caption*{(a) $\alpha_v$ on CUB}
    \end{minipage}
    \hfill
    \begin{minipage}{0.48\linewidth}
        \centering
        \includegraphics[width=\linewidth]{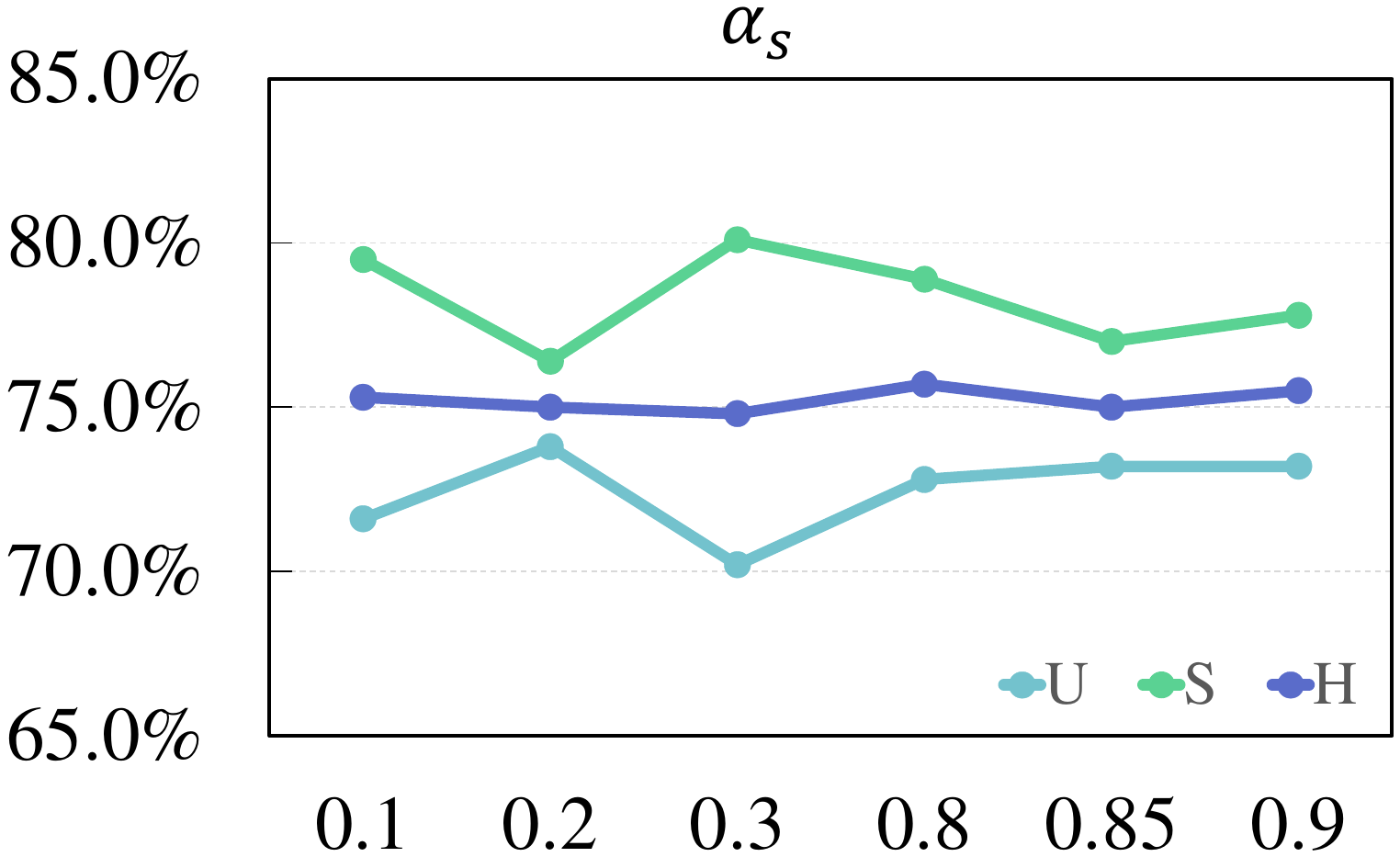}
        \captionsetup{labelformat=empty, skip=4pt}  
        \caption*{(b) $\alpha_s$ on CUB}
    \end{minipage}
\vspace{-1.0em} 
\caption{The impact of hyperparameters $\alpha_v$ and $\alpha_s$ to control the importance of self-attention and bias in the strong fusion process.}
\label{fig:av_as}
\vspace{-1.0em} 
\end{figure}

To give further insight into each component in our framework, we perform ablations to analyze the effectiveness of our VSPCN, as shown in Table \ref{table:results ablation}. We conducted ablation experiments focusing on three key aspects: (1) visual and semantic prompts; (2) weak and strong fusion; (3) semantic adapter. We will discuss them in detail.

\noindent
\textbf{Effectiveness of visual and semantic prompts.} In the VSPCN, we use the ViT as the baseline to extract visual features and calculate the similarity score with category prototypes. We also show the results of using visual and semantic prompts separately. Specifically, the model with visual prompt achieved a gain on $H$ by (13.4\%, 6.2\%, 3.6\%) on (CUB, SUN, AWA2) compared with the baseline, while the model with semantic prompt achieved a higher improvement by (14.6\%, 7.0\%, 11.2\%). This indicates that the semantic prompt plays an important role in refining the semantic-related visual features. With the collaboration of visual and semantic prompts, the performance is improved.

\noindent
\textbf{Effectiveness of weak and strong fusion.} We explore three situations: prompts without fusion, prompts with weak fusion, and prompts with strong fusion. The results in Table \ref{table:results ablation} (lines 3-6) indicate that both weak and strong fusions are necessary to achieve effective feature adaption.

\noindent
\textbf{Effectiveness of semantic adapter.} We utilize adapters to adjust the semantic features. To demonstrate their effectiveness, we experiment without adapters, and the results are shown in the seventh line of Table \ref{table:results ablation}. It can be seen that the performance drops without adapters, which indicates that it is necessary to learn adaptive semantic features.

\subsection{HyperParameter Analysis}

We perform experiments to investigate the effects of various hyper-parameters on CUB and SUN datasets, including $\lambda_{CED}$ and $\lambda_{SKD}$ in Eq. (\ref{Eq:loss}), $\alpha_v$ in Eq. (\ref{Eq:alpha_v}) and $\alpha_s$ in the Eq. (\ref{Eq:alpha_s}). From the results in Fig. \ref{fig:CED_SKD} (a), with the increase of $\lambda_{CED}$, the harmonic mean raised slowly at first and then decreases when $\lambda_{CED}>0.8$ on the CUB dataset. Larger $\lambda_{CED}$ will enhance the visual prompt for discriminative feature learning. Similarly, as shown in Fig. \ref{fig:CED_SKD} (b), VSPCN achieves better performances when the value is set as $\lambda_{CED}=0.4$ on SUN. From Fig. \ref{fig:CED_SKD} (c) and (d), it can be seen that as $\lambda_{SKD}$ changes, $U$ and $S$ fluctuate significantly; hence, we choose $\lambda_{SKD}$=0.9 on CUB and $\lambda_{SKD}$=0.35 on SUN to obtain balanced performance between $U$ and $S$.

The parameters $\alpha_v$ and $\alpha_s$ adjust the weight between the attention score and the bias score, which are utilized to fuse the prompt information. As shown in Fig.\ref{fig:av_as} (a)(b), the optimal values for  $\alpha_v$ and $\alpha_s$ are 0.05 and 0.8, which reflect that visual prompt focuses on attention score, while semantic prompt focuses on the bias scores on the CUB dataset.

\begin{figure*}[t]
\centering
\includegraphics[width=1.0\linewidth]{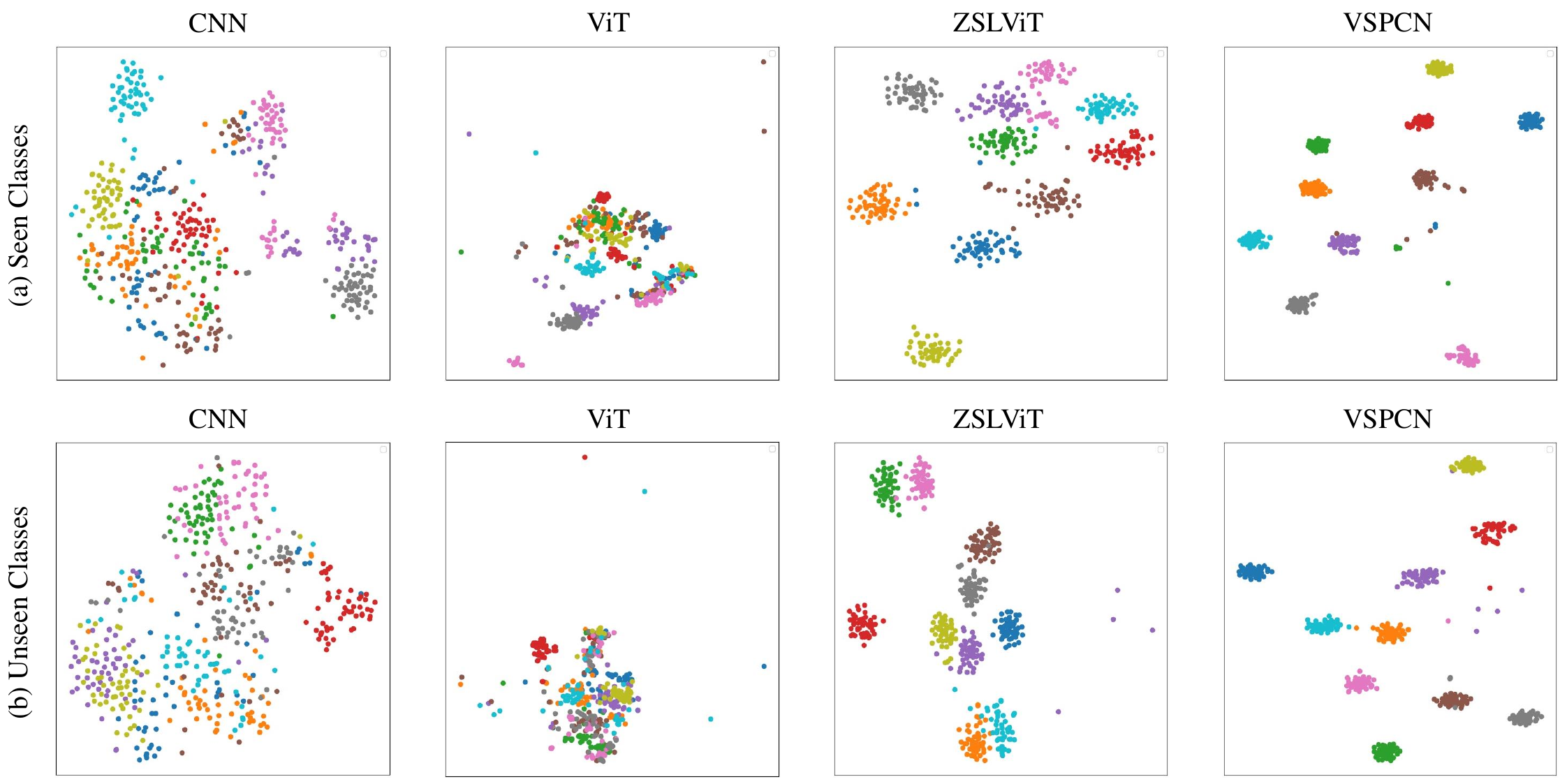}
\vspace{-1.0em}
\caption{ Feature visualization for seen classes and unseen classes on CUB by t-SNE. Different colors refer to different classes. We randomly select 10 classes and show the visualization results of different approaches. }
\label{fig:tSNE}
\vspace{-1.0em}
\end{figure*}

\begin{figure}[h]
\centering
\includegraphics[width=0.95\linewidth]{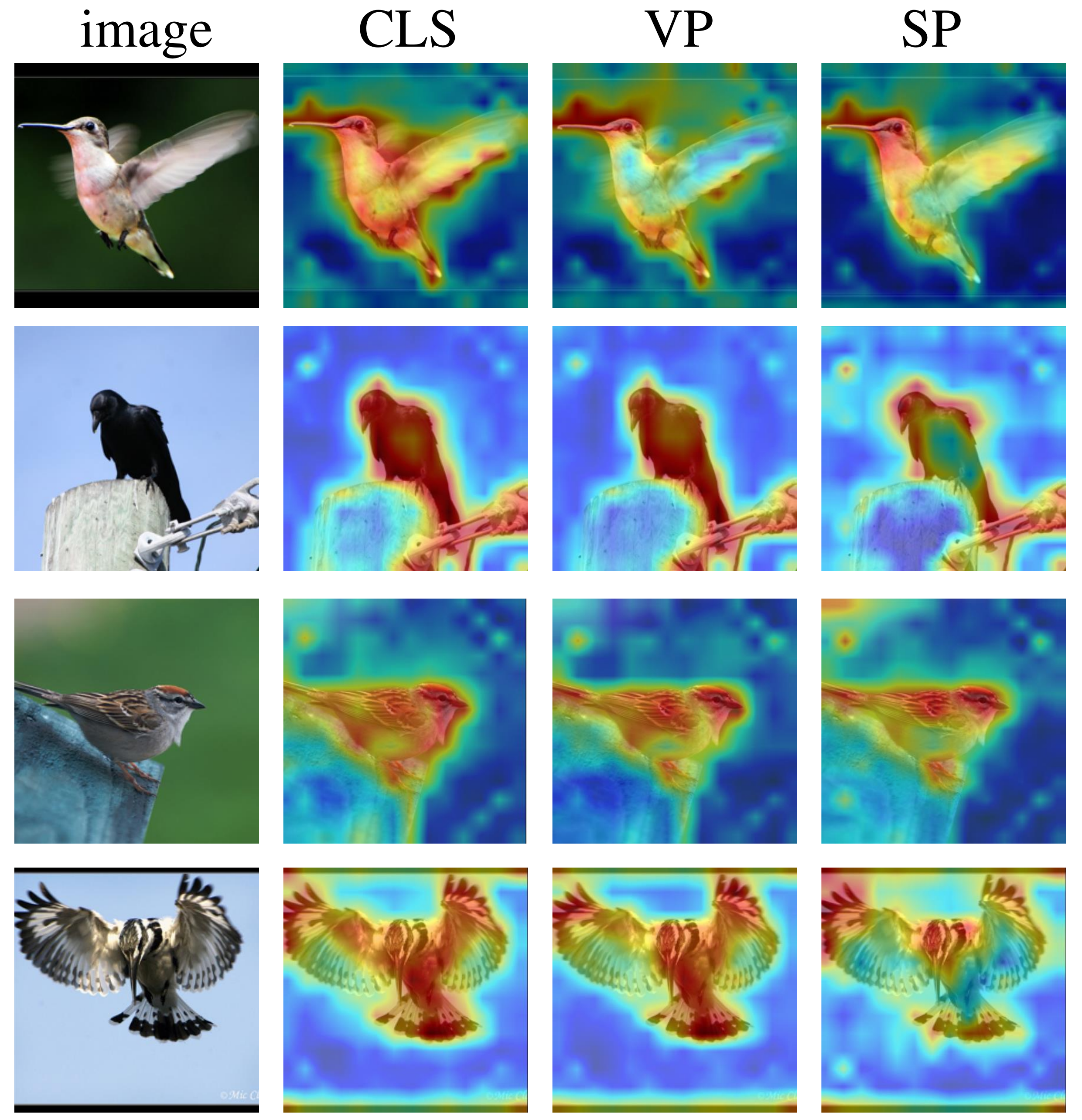}
\vspace{-1.0em} 
\caption{Visualization of attention maps obtained by CLS token, visual prompt (VP), and semantic prompt (SP), respectively.}
\label{fig:maps}
\vspace{-2.0em} 
\end{figure}

\subsection{Qualitative Results}

\noindent
\textbf{Visualization of Visual Features.}
To gain an intuitive understanding of the visual features, we visualize the visual features extracted by our VSPCN using t-SNE \cite{JMLR:v9:vandermaaten08a}. As shown in Fig \ref{fig:tSNE}, we compare our approach with other approaches, including 
 the CNN backbone (ResNet101 \cite{he2016deep}), ViT-Base \cite{dosovitskiy2020image}, and ZSLViT \cite{chen2024progressive}. We choose 10 seen classes and 10 unseen classes for visualization, where 50 samples are randomly selected for each class. It can be seen that compared to ZSLViT and VSPCN, the features extracted directly using pre-trained CNN and ViT are less discriminative for different classes, so it is necessary to adapt the general model to specific tasks. Furthermore, comparing the results of ZSLViT and VSPCN, we can see that the visual features learned by VSPCN achieve better performance for intra-class compactness and inter-class separability, leading the classifier to learn appropriate classification boundaries. 

 \noindent
\textbf{Visualization of Attention maps.}
To intuitively demonstrate the effectiveness of our VSPCN framework, we visualized the attention maps of some samples on the CUB dataset. Specifically, we perform attention calculations on the CLS token, visual prompt, and semantic prompt respectively, and the attention maps are shown in Fig.\ref{fig:maps}. We observe that both the visual prompt and semantic prompt are capable of identifying the key regions of objects. However, these regions represent only a portion of the object. In contrast, the CLS token (CLS) integrates information from both the visual prompt and the semantic prompt, which produces a more comprehensive attention map. Therefore, we use the CLS token for recognition.

%% file: 05_conclusion.tex
\subsection{Conclusion}
In this paper, we propose a Visual and Semantic Prompt Collaboration Network (VSPCN) for GZSL, where dual prompts complement each other to effectively adapt the pre-trained ViT model for semantic-relevant visual feature learning. Specifically, the visual prompt integrates the image features for discriminative information extraction and the semantic prompt integrates the semantic information for semantic information extraction. To achieve effective information integration, we further design weak and strong prompt fusion mechanisms for different layers in our model. Moreover, we design adapters to learn adaptive semantic features to enhance the semantic information. Through the collaboration of visual and semantic prompts, our framework can learn adaptive visual features that generalize well for novel categories. Extensive experiments on three benchmark GZSL datasets show the effectiveness of the proposed framework.

%% file: Acknowledge.tex
\noindent
\\ \\
\textbf{Acknowledgements.} This work was supported by National Key R\&D Program of China No.2021ZD0111902 and National Natural Science Foundation of China under Grant No.62206007 and  Grant U21B2038. Xiaohan Yu, Jian Yang and Yuankai Qi are not supported by the aforementioned fundings.

%% file: 12_appendix.tex
\appendix
\label{sec:appendix}
